\documentclass{article}

\usepackage[preprint]{corl_2026} % Uncomment for pre-prints (e.g., arxiv); This is like ``final'', but will remove the CORL footnote.

\usepackage{amsmath}
\usepackage{algorithm, algpseudocode}
\usepackage{graphicx}
\usepackage{booktabs, tabularx}

\title{Program-Level Adversarial Co-Adaptive Self-Play for Robotic Security}

\author{
  Leo Benaharon\thanks{Equal contribution.} \thanks{The authors are with Isuzu Technical Center of America, Plymouth, MI 48170 USA.
  {lbenahar@purdue.edu, \{chi.zhang, wei.zheng, weifan.zhang, yifan.wei, yong.sun\}@isuzu.com}} \\
  \And
  Chi Zhang\footnotemark[1] \footnotemark[2] \\
  \And
  Wei Zheng\footnotemark[2] \\
  \And
  Weifan Zhang\footnotemark[2] \\
  \And
  Yifan Wei\footnotemark[2] \\
  \And
  Yong Sun\footnotemark[2] \\
}

\begin{document}
\maketitle

%===============================================================================

\begin{abstract}
    % Task completion under attack does not by itself establish effective robotic defense. We present a framework for program-level adversarial co-adaptation, in which coding agents revise executable attacks and defenses using tested opponent artifacts and mission feedback without model fine-tuning. Environment-specific contracts support external autonomy-stack programs and robot-side SDK modifications. Two eight-generation case studies examine distinct configurations. The driving study records 18 goal completions, 16 timeouts, and 6 collisions, including a traced transition to direct-actuation recovery. The humanoid study exercises orchestrated attacker allocation, development-time sparring, and clean, attack-only, defense-only, and repeated joint evaluations. Across 176 formal episodes, 30 of 40 attacker-generation comparisons are eligible for scoring; four attain median normalized recovery of 1.0, while worst-case recovery remains at most 0.0301. One such aggregate includes a fall among its three repeats. These observations reveal local program responses without broad defensive improvement and show why attack validity, repeat-level outcomes, and changing opponents must accompany aggregate scores.

    Coding agents can adapt robotic software by revising executable programs, but successful responses to an evolving opponent may not remain effective. We present a framework for program-level adversarial co-adaptive self-play in which coding agents generate attacks and defenses from tested opponent code and mission feedback, without model fine-tuning. Environment-specific interfaces support external programs for autonomous driving and SDK-level defenses for humanoid locomotion. 
    % The humanoid configuration adds attacker orchestration, development-time testing, and four-condition evaluation that separates attack effectiveness from defensive recovery. 
    Two eight-generation case studies reveal concrete software responses: a driving defender introduces direct-actuation recovery, while a humanoid defender revises incoming-state validation. These responses provide local task recovery but do not establish sustained robustness. 
    % In the humanoid study, a maximum median recovery still includes a failed execution, and worst-case recovery remains low. 
    The framework makes such adaptation and limitations observable through executable artifacts and task-level evidence.
    
    % Coding agents can modify robotic behavior through executable software rather than updates to model parameters. To investigate this form of adaptation in a security setting, we present a framework for program-level adversarial co-adaptation in which fixed-weight coding agents generate and iteratively revise attack and defense programs around an autonomous-driving stack. Five specialist attackers and one shared defender interact with Autoware through restricted ROS 2 interfaces in CARLA. Each generation freezes the submitted programs, evaluates mission outcomes using an independent referee, and returns tested opponent code and execution feedback for subsequent revision. In an eight-generation case study on a single fixed driving scenario, the defender succeeds in all five matches at generation three, followed by renewed failures, particularly on control and chained interfaces. These preliminary observations highlight why successful responses to current opponents do not establish sustained robustness. The framework provides an executable setting for studying adversarial program adaptation in robotic systems.
\end{abstract}


%===============================================================================

\section{Introduction}
\label{sec:introduction}
	Robotic cybersecurity must protect physical behavior as well as software access. A disclosed Unitree Go2 vulnerability exposed a DDS programming interface without authentication or authorization, permitting a network-adjacent attacker to install code later executed through a controller binding~\cite{unitreeCVE}. Such risks motivate proactive testing: develop attacks and candidate defenses before malicious exploitation, then evaluate their consequences for an ongoing robotic task.

Coding agents provide a way to study this interaction through executable programs. Prior work connects generated code to embodied control \citep{liang2022code}, develops executable skills through environmental feedback \citep{wang2023voyager}, and investigates code-mediated improvement and autonomous cyber co-evolution \citep{robeyns2025self,meng2026sysevolve}. In this work, we study a system-level
question: \textbf{can agents attack an existing robotic stack and revise defenses effectively as
the opponent adapts?}

Our framework implements \emph{program-level adversarial co-adaptive self-play}. Attack-generating agents and a shared defense-generating agent are placed around a running robotic system. Programs submitted by agents are checked and frozen, executed on the robotic system, and returned with independently observed outcomes to inform subsequent revisions. Adaptation occurs in software and retained context rather than through arena-driven model fine-tuning. Environment-specific tasks and permissions determine what those programs may change.

We study this process in Autoware-CARLA and Unitree G1-MuJoCo. The driving configuration uses fixed attacker roles; the humanoid configuration additionally uses an orchestrator, development-time tests against frozen opponents, and separate clean, attack-only, defense-only, and joint episode executions. The studies reveal concrete program revisions alongside recurrent failures, motivating evaluation beyond a favorable individual match.

Our contributions are:

\begin{itemize}
\item
  \textbf{A robotic-security self-play framework with directly competing programs}. Independent coding agents revise programs that execute together around an operating autonomy stack through robotic middleware and permitted software modifications.
\item
  \textbf{A mission-grounded evaluation protocol.} The humanoid evaluation combines clean, attack-only, defense-only, and repeated joint trials to distinguish attack effectiveness, defense viability, and normalized task recovery.
\item
  \textbf{Two eight-generation case studies of adversarial adaptation.} Driving and humanoid experiments trace how agents develop countermeasures and counterattacks in response to each other, revealing targeted recovery alongside persistent failures.
\end{itemize}

%===============================================================================

% We may drop this section
\section{Related Work}
\label{sec:relatedwork}
    \textbf{From executable behavior to competitive adaptation.} Code as Policies connects language-model-generated programs to robotic perception and control interfaces \citep{liang2022code}. Voyager extends feedback-driven program development to a growing library of Minecraft skills without model fine-tuning \citep{wang2023voyager}. 
% Competition introduces a further challenge: useful behavior must adapt to an opponent that also changes. 
CodeClash alternates independent code revision with competition between the resulting programs, and separately studies access to previous opponent code \citep{yang2025codeclash}. FAMOU evolves strategy programs against an updating opponent pool \citep{li2026famou}, while COvolve jointly generates policies and challenging environments, including CARLA scenes \citep{sygkounas2026covolve}. Our experiments keep the mission and simulator configuration fixed and instead revise attack and defense programs around an existing robotic stack.

\textbf{From competition to security hardening.} Cybersecurity gives the two sides different responsibilities: disrupting a system and preserving useful operation. Hemberg et al.\ study cyber-controller coevolution, including LLM-assisted mutation of Python controllers \citep{hemberg2025coevolution}. 
% Their evolutionary search is a precedent for executable cyber coadaptation; 
Our agents instead revise programs through coding sessions without population-level selection. RvB couples vulnerability discovery with report-driven patching and verification \citep{huang2026rvb}, while ZERO-APT evaluates attack execution against a log-based detection agent in an attacker--defender--judge loop \citep{zheng2026zeroapt}. We study generated attack and defense programs that compete directly during a robotic mission, with the outcomes guiding both agents’ next program revisions.

\textbf{From software interference to robot behavior.} RoboFuzz uses ROS-aware mutation and correctness oracles to detect semantic failures, connecting software tests to robotic behavior rather than crashes alone \citep{kim2022robofuzz}. Somma et al.\ apply an environment-grounded multi-agent workflow to ROS/ROS~2 penetration testing \citep{somma2026environment}. Haque et al.\ investigate LLM-generated dynamic tests for Autoware \citep{haque2026dynamic}.
% build integration prevents confirmation of candidate weaknesses within their reported budget
RoboPAIR studies a different entry point, adversarial instructions to LLM-controlled robots \citep{robey2024robopair}. Here the language models develop the attack and defense software; the targeted autonomy stack need not itself contain an LLM. Combining this software-level access with mission outcomes lets us examine not only whether interference succeeds, but how generated countermeasures alter the robot's behavior and become targets of later attacks.

%===============================================================================

\section{Methodology}
\label{sec:methodology}
    % Figure~\ref{fig:framework} summarizes the shared loop and configuration-specific components. Program development, including sparring where enabled, precedes formal evaluation of frozen artifacts. Environment adapters specify tasks, interfaces, submission formats, resets, and outcome definitions. The original driving study predates the adapter refactor and does not validate its later live Autoware path.

\begin{figure}[htbp]
    \centering
    \includegraphics[width=0.8\linewidth]{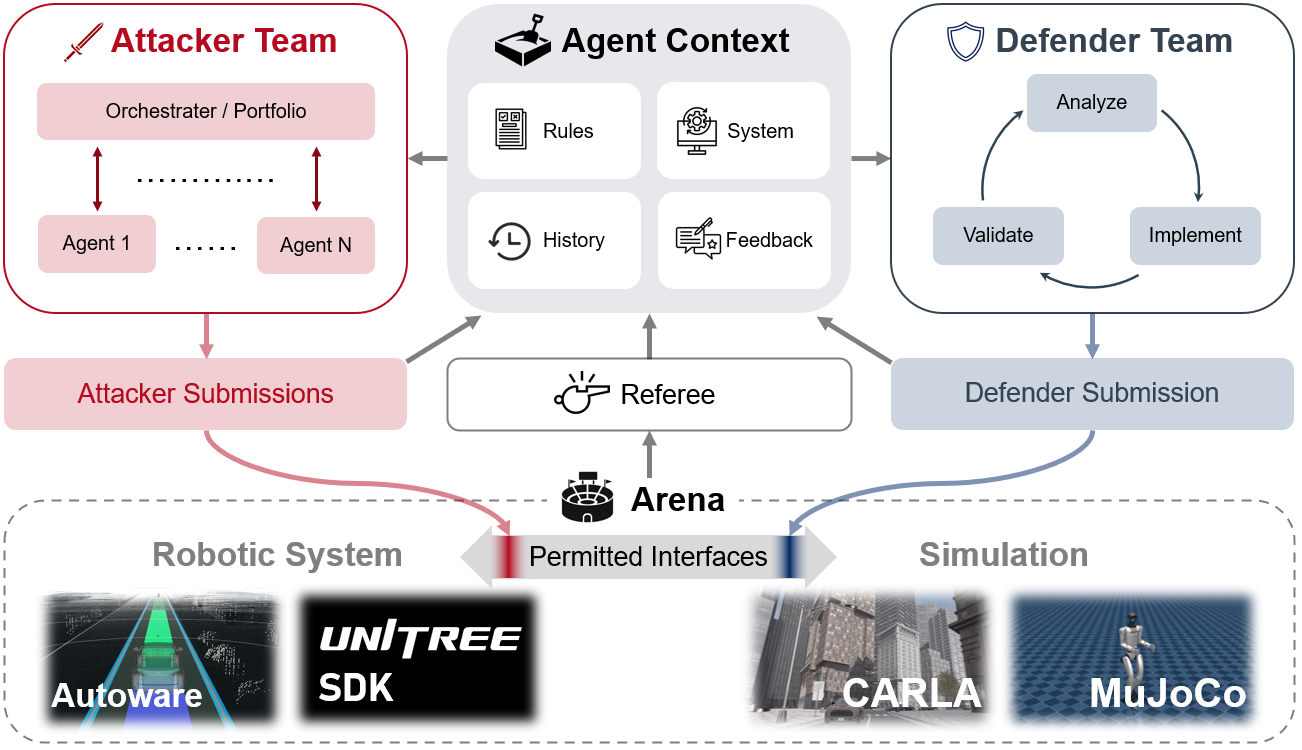}
    \caption{Program-level adversarial co-adaptation. Coding agents develop attack and defense programs that are checked, frozen, and evaluated in Autoware-CARLA or G1-MuJoCo. Independent referees judge outcomes from simulator ground truth; results and tested opponent code guide the next generation.
    }
    \label{fig:framework}
\end{figure}

Figure~\ref{fig:framework} summarizes the
program-generation and evaluation loop. A fixed mission focuses adaptation on the opponent's software. One shared defender faces each attacker separately, so it must accommodate different forms of interference with the same submission.

\subsection{Attack-Defense Self-Play Loop} \label{program-level-adversarial-interaction}

Before the self-play loop begins, the arena verifies a clean mission as baseline and captures exposed interfaces for agent context. At each generation, attacker (Red) and defender (Blue) agents develop concurrently in private workspaces. Each receives role-specific context describing the rules, robotic software, retained history, and previous execution feedback. Each Red sees its own preceding evidence and the tested defender; Blue receives the preceding attackers and their match records. Current development workspaces are hidden from opponents.

After development, the arena checks submissions and permits up to two automated repairs. Middleware access is assumed, and direct manipulation of the simulator or referee is outside the interaction policy.

Environment-specific permissions define the allowed interventions. For driving, both sides use external ROS 2 programs, and Blue may issue direct actuation commands. For the humanoid, Red uses DDS messages, while Blue supplies a monitor and permitted SDK replacements that process inputs before the fixed locomotion policy consumes them. The monitor can also publish commands.

Accepted programs are checked, frozen, and evaluated, with one attacker paired with the same generation-specific defense in each joint episode. They execute without live language-model calls or mid-trial edits. An independent referee determines outcomes from simulator ground truth while the results and tested opponent code are returned for subsequent revision. The robot resets between trials.

\subsection{Attacker Allocation and Evaluation}\label{attacker-allocation-and-Evaluation}

The driving configuration assigns five fixed specializations: perception, failsafe, planning, control, and chained attacks. In the humanoid configuration, an orchestrator proposes five roles before each generation using the available interfaces, task, and prior feedback. Role identities preserve their histories; unused attackers remain available to Blue. The orchestrator is instructed to retain effective pressure and explore other directions when roles become ineffective or well recovered.

% Humanoid agents may also run bounded development-time tests, termed \emph{sparring}, either alone or against frozen past opponents. These tests return execution evidence for program revision but are excluded from formal evaluation. The distinction prevents exploratory testing from being counted as independent evidence of final performance.

Shorter humanoid episodes and up to sixteen parallel slots make more testing practical than the sequential driving setup with per-match Autoware restarts. This supports both unscored development sparring, where agents test candidate programs alone or against frozen prior opponents, and more formal evaluation episodes. Each generation runs one clean, one defense-only, five attack-only, and fifteen joint episodes (three per attacker): $1+1+5+15=22$ formal evaluations.

\subsection{Mission-Grounded
Evaluation}\label{mission-grounded-evaluation}

\textbf{Driving reward and revision feedback.} Blue's terminal reward is $+1$ for reaching the goal, $-1$ for collision or off-road departure, and $0$ for timeout or no progress; Red receives its negative. Timeouts are draws regardless of how far the car travels. Terminal rewards do not capture partial mission recovery, so a separate quality score exposes differences in progress and speed for revision.

Let $L$ be the start-to-goal distance and $d_{\max}$ the greatest recorded progress, giving the route fraction $p=\text{clip}_{[0,1]}(d_{\max}/L)$, set to one on goal completion. Let $v_{\max}$ be peak vehicle speed and $v_{\mathrm{ref}}$ the clean-baseline peak, giving $a=\text{clip}_{[0,1]}(v_{\max}/v_{\mathrm{ref}})$. Driving quality is $q_D=p(0.5+0.5a)$, with $L=60$ m and $v_{\mathrm{ref}}=4.5$ m/s. Progress rewards task completion, while retained speed discourages a slow defensive takeover and encourages preserving Autoware's nominal driving behavior. This is a lightweight proxy. \emph{Quality informs revision but is not added to the terminal reward.}

\textbf{Humanoid quality and reward.}
A faster fall should not outscore a completed walk, so humanoid quality separates partial progress from completion:
\begin{equation}
q_H=
\begin{cases}
0.5+0.5a,&\text{goal reached},\\
0.5p,&\text{otherwise}.
\end{cases}
\label{eq:quality}
\end{equation}
Here, $p$ is progress along the 3 m route and $a$ is peak body walking speed relative to the clean baseline. Failed episodes are rewarded for progress; speed distinguishes successful walks.

Unlike driving, humanoid evaluation includes attack-only and repeated joint episodes, allowing reward to measure how much attack-induced loss Blue restores. Let $q_C$ be the quality in a clean episode, $q_{A,k}$ be the quality in an attack-only episode, and $q_{J,k,j}$ quality in joint episode $j$:
\begin{equation}
\rho_{k,j}=\text{clip}_{[0,1]}
\left(\frac{q_{J,k,j}-q_{A,k}}{q_C-q_{A,k}}\right).
\end{equation}
Thus, $\rho=0$ means no improvement over attack-only and $\rho=1$ restores clean quality. For gaps below 0.05, recovery falls back to goal completion: 1 if reached and 0 otherwise.

Blue's reward is the median of the three joint-episode recovery scores; Red receives its complement. The median limits the effect of one lucky success or one failed episode. A matchup is eligible when the attack disrupts the undefended mission, the defense preserves the defense-only mission, and the joint episodes are valid. Blue's lowest eligible reward guides revision toward its weakest defense.

%===============================================================================

\section{Experimental Results}
\label{sec:result}
    \subsection{Experimental Setup}\label{experimental-setup}
Table~\ref{tab:setup} summarizes the two robotic platforms, agent configurations, attacker allocation, and evaluation settings. Across both studies, we trace implemented attack and defense revisions and link them to their resulting mission outcomes.
% We trace attacks, defensive responses, and subsequent outcomes, distinguishing implemented revisions from recorded strategy proposals. Table~\ref{tab:setup} summarizes the configurations. 
% Additional task parameters and available model information are provided in Appendix~\ref{app:details}.

\begin{table}[!htbp]
\centering
\caption{Evaluated configurations. Each uses one eight-generation run and one shared defender per generation.}
\label{tab:setup}
\small
\renewcommand{\arraystretch}{1.12}
\begin{tabularx}{\linewidth}{@{}p{0.23\linewidth}XX@{}}
\toprule
 & Driving & Humanoid \\
\midrule
Task & 60 m clear-road route & 3 m flat-ground walk \\
Robot software & Autoware in CARLA & G1 policy + navigation in \mbox{MuJoCo} \\
Coding harness & Claude Code / Opus & Claude Code / Sonnet \\
Agent budget & 40 min per session & 60 min per session \\
Attacker roles & Five fixed & Five orchestrated \\
Joint episode repetitions & One per attacker & Three per attacker \\
Additional controls & Initial clean trial & Clean, defense only, attack only \\
Execution & Sequential; stack reset & Up to 16 isolated parallel slots \\
\bottomrule
\end{tabularx}
\end{table}

\subsection{Driving}\label{driving-task-recovery-and-renewed-failures}

\begin{figure}[htbp]
    \centering
    \includegraphics[width=1.0\linewidth]{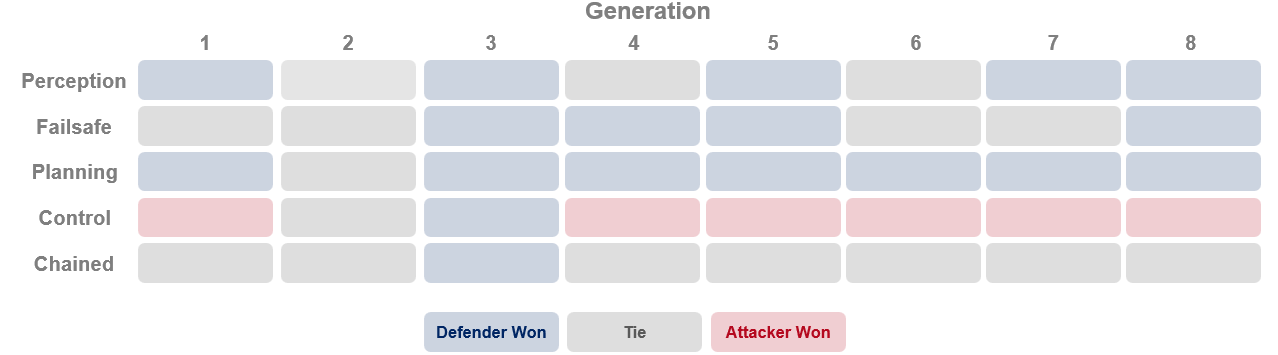}
    \caption{\textbf{Driving.} A crash, collision, or off-road event results in an attacker win; successful completion results in a defender win; and a timeout results in a tie.
    }
    \label{fig:driving_gens}
\end{figure}

Across 40 driving matches, 18 reach the goal, 16 time out, and 6 end in collision. The evaluated route contains no other traffic.
Figure~\ref{fig:driving_gens} summarizes these outcomes, while the progression below highlights how wins and losses shift across generations as new attacks defeat prior defenses, countermeasures regain ground, and some failures persist despite adaptation.  

\begin{figure}[htbp]
    \centering
    \includegraphics[width=0.9\linewidth]{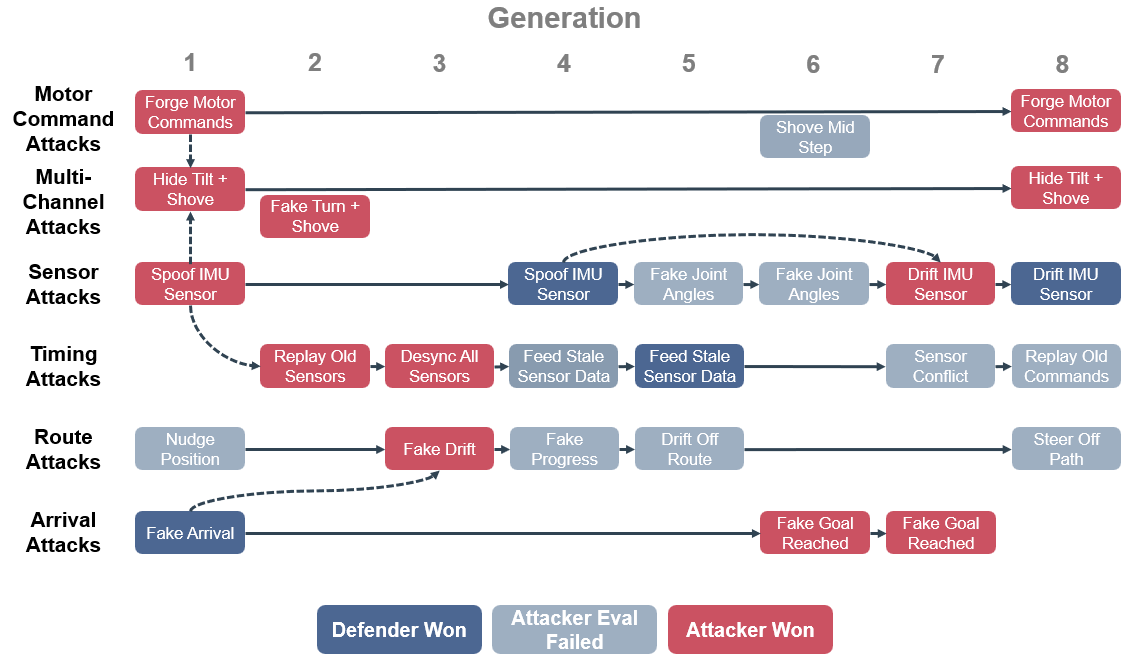}
    \caption{\textbf{Humanoid.} Each box summarizes an orchestrator-assigned task, grouped by attack type. Solid arrows show continuation of a strategy while dashed arrows show borrowed mechanisms. Blue marks eligible matchups where the median of the three joint-episode recovery scores is 1; red marks the remaining eligible matchups. Gray marks attacks that do not disrupt attack-only execution and are excluded from scoring.
    }
    \label{fig:mujoco_summary}
\end{figure}

% \begin{table}[t]
%     \centering
%     \caption{Driving terminal outcomes. G: goal reached; T: timeout; C: collision. Each cell evaluates one attacker against the shared defender for that generation.}
%     \label{tab:2}
%     \begin{tabular}{@{}cccccc@{}}
%         \toprule
%         Generation & Perception & Failsafe & Planning & Control & Chained \\
%         \midrule
%         1 & G & T & G & C & T \\
%         2 & T & T & T & T & T \\
%         3 & G & G & G & G & G \\
%         4 & T & G & G & C & T \\
%         5 & G & G & G & C & T \\
%         6 & T & T & G & C & T \\
%         7 & G & T & G & C & T \\
%         8 & G & G & G & C & T \\
%         \bottomrule
%     \end{tabular}
% \end{table}

\textbf{Generations 1-3:} Initial message repairs do not prevent five timeouts in generation 2. Blue identifies a common downstream effect across false objects, zero speed limits, and stop requests. Generation 3 introduces a recovery driver at their shared actuation interface, and all five matches reach the goal.

\textbf{Generation 4:} The recovery driver disables its bypass when it detects another publisher on \texttt{/control/command/actuation\_cmd}. The control attacker deliberately publishes there, triggering the stand-down; the match ends in collision.

\textbf{Generations 5-6:} Red shifts toward targeting the new defense itself. Perception targets the recovery driver's LiDAR inputs, Control increases command publication, and Chained competes with Blue's throttle through braking. Blue responds with map and lane based recovery and dedicated command handling. In generation 6, Failsafe exploits apparent goal arrival, producing a timeout.

\textbf{Generations 7-8:} Blue adds arrival cross-checks, stall detection, lane-center recovery, and counter-steering. Red responds with intermittent false obstacles and incorrect-path motion intended to evade these checks. Perception and Planning reach the goal in both generations, while Failsafe recovers by generation 8. 

\begin{figure}[htbp]
    \centering
    \includegraphics[width=1.0\linewidth]{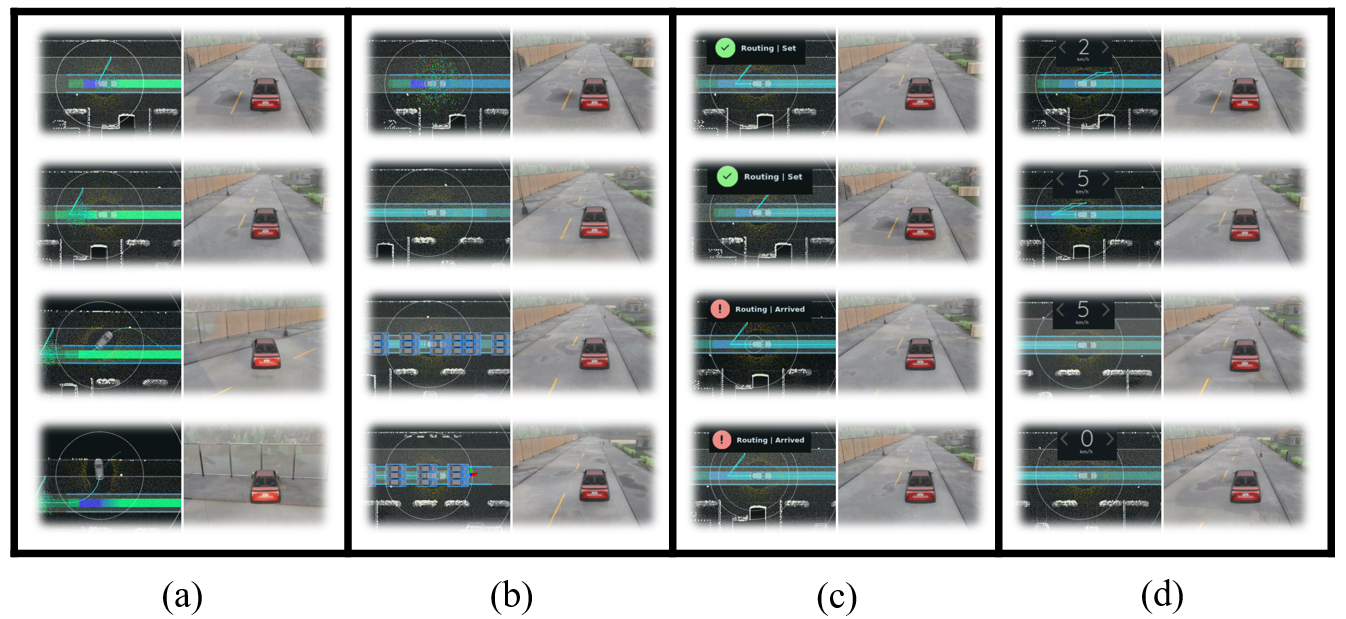}
    \caption{\textbf{Driving.} Each column progresses from top to bottom, pairing the Autoware visualization (left) with the simulated vehicle (right). (a) the attacker overwhelms the defender with competing steering commands; (b) the defender mitigates fabricated vehicle detections; (c) a false arrival signal prevents mission completion; (d) the attacker progressively reduces the speed limit to 0 km/h.
    }
    \label{fig:rollout_driving}
\end{figure}

Figure~\ref{fig:rollout_driving} shows representative cases where adaptation succeeds and where attacks continue to break the defense. The humanoid study provides a complementary view, using attack-only controls and repeated joint episodes to separate attack effectiveness from defensive recovery.

\subsection{Humanoid}\label{humanoid-local-recovery-under-persistent-pressure}

Across 176 formal humanoid evaluations (8 clean, 8 defense-only, 40 attack-only, and 120 joint), 51 reach the goal. All clean and defense-only episodes succeed, while 30 of 40 attack-only episodes disrupt the mission. Of the 120 joint episodes, 25 reach the goal, including 11 of 90 corresponding to attacks that disrupt attack-only execution. Agents also run 253 development tests, 57 against frozen opponents.
Figure~\ref{fig:mujoco_summary} traces how the orchestrator's attack assignments evolve across generations alongside their outcomes against the defender. The following analysis examines how attacks and defenses change in response to earlier outcomes.

\textbf{Generations 1-3:} Blue initially checks message order and position plausibility. Its SDK notes expose the next weakness: old state can appear newer, while small position changes can accumulate over time. Blue adds orientation checks and bounds on tick advancement and cumulative drift. Nevertheless, generation 3 produces no joint episode that reaches the goal.

\textbf{Generations 4-5:} Blue's earlier filter rejects large tilt readings, yet IMU spoofing still causes falls. Its notes identify the gap: early values in Red's gradual false-tilt ramp still pass the filter. In generation 4, Blue lowers the accepted IMU tilt from 0.35 to 0.22 rad; two of three joint episodes then reach the goal. The orchestrator retires IMU spoofing but retains the command attacks, while new joint-state and route-drift roles target other inputs.

\textbf{Generations 6-8:} The goal-spoofing role introduced in generation 6 targets the filter's initial position reference. Generation 7 adds a bound referenced to the known start and rejects samples on validation errors. A later goal-spoofing attack instead ramps reported progress below the existing speed limit, while a new IMU-drift role keeps angular-velocity bias below Blue's threshold. Blue responds in generation 8 by tightening both checks. 

\begin{figure}[htbp]
    \centering
    \includegraphics[width=.7\linewidth]{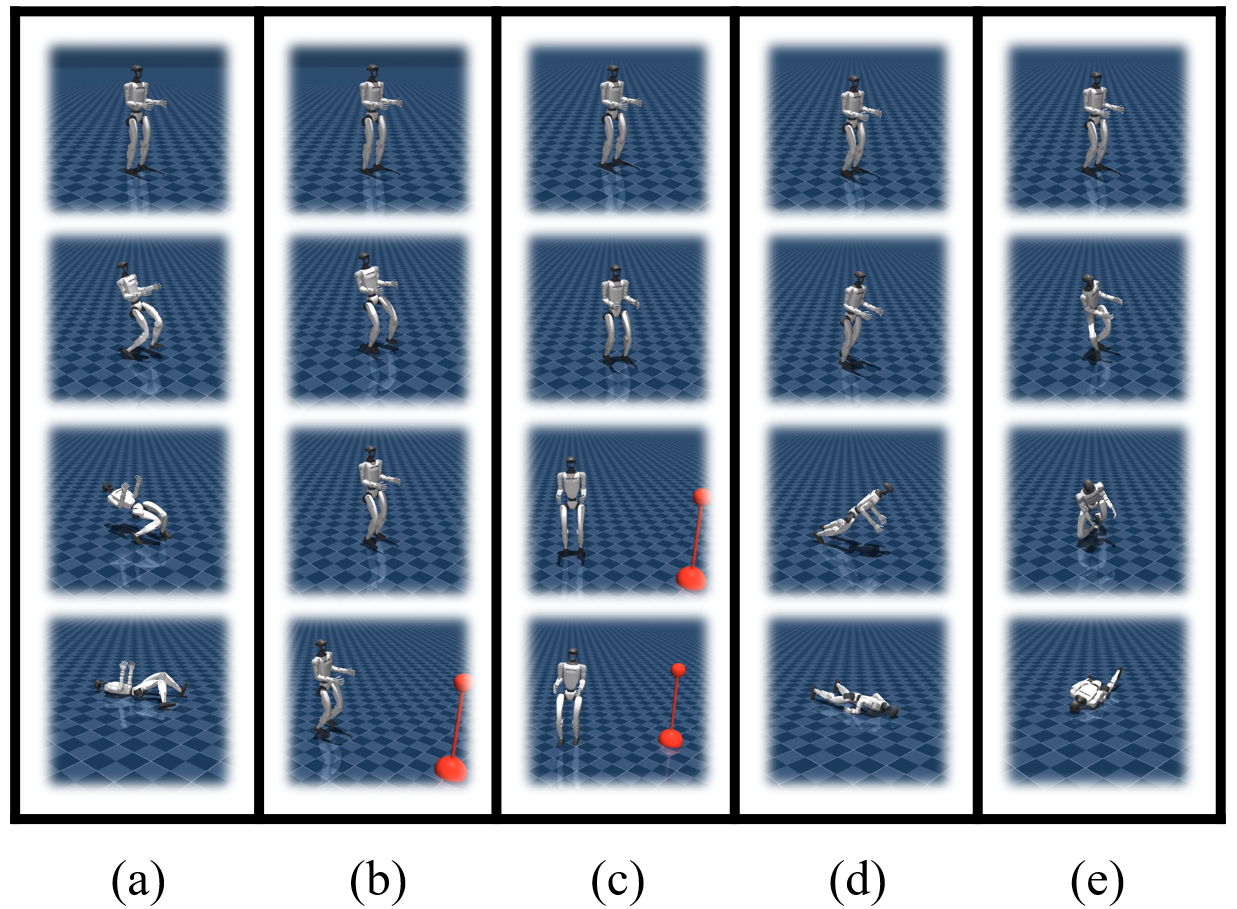}
    \caption{\textbf{Humanoid.} Each column progresses from top to bottom. The red marker indicates the goal. (a) IMU spoofing causes a fall without defense; (b) the defender mitigates the same attack; (c) Position drift sends the robot off-route; (d) Combined sensor spoofing and motor-command injection and (e) competing actuator commands both cause falls.
    }
    \label{fig:rollout_humaniod}
\end{figure}

Figure~\ref{fig:rollout_humaniod} visualizes this progression
through representative rollouts. Across both studies, agents repeatedly respond to each other's revisions: defenses recover specific failures, later counterattacks target those defenses, and some attack classes remain unresolved.

%===============================================================================

\section{Limitations and Future Work}
\label{sec:limitations}
    % \textbf{Scenario robustness and causal scope.} Each configuration has one eight-generation lineage and a simple task. To establish broad robustness and isolate each revision’s effect, future studies should evaluate retained opponents under matched budgets and broaden task conditions.

\textbf{Scenario robustness.} Each configuration has one eight-generation lineage and a simple task. To establish broad robustness, future studies should broaden scenario and task conditions.

% \textbf{Evaluation metrics and defensive authority.} Both quality functions use peak speed and progress rather than comprehensive safety, smoothness, or control provenance. Their normalized values are configuration-specific. Clipping and median aggregation can obscure residual failures. 
% Driving defenses may substitute direct control, while humanoid SDK filters have different authority over inbound state and competing commands. These observations do not certify subsystem protection or hardware security.

\textbf{Evaluation metrics.} Both quality functions use peak speed and progress rather than comprehensive safety, smoothness, or control provenance. Their normalized values are configuration-specific. Clipping and median aggregation can obscure residual failures. 

\textbf{Agent ablations.} We do not isolate which feedback mechanisms most improve agent learning. Future work should ablate tested opponent code, prior-match history, development sparring, and continuous quality feedback to determine which most improve adaptation across generations.

\textbf{Future defender orchestrator.} A future defender orchestrator could distribute development among specialists, but their changes must share SDK hooks, state, and command authority. Merge checks, rollback, and tests against historical attacks are needed to add protections without removing or conflicting with earlier ones.
% A future defender orchestrator could distribute development across specialized agents, each targeting different attack surfaces. Their defenses, however, must share SDK hooks, state, and command authority without interfering with one another. The orchestrator should therefore enforce merge checks, rollback mechanisms, and regression tests against historical attacks so that new protections do not remove, weaken, or conflict with previously effective defenses. Evaluating this design under matched budgets and fixed historical opponents would help determine whether coordinated specialization improves defensive robustness over a single defender.

%===============================================================================

\section{Conclusion}
\label{sec:conclusion}
    We present a framework for program-level adversarial co-adaptive self-play using executable robotic software, opponent artifacts, and mission feedback. Two eight-generation studies trace direct task recovery and SDK-based defensive responses while revealing recurrent failures. 
% Four-condition evaluation and repeat-level reporting distinguish local recovery from ineffective attacks and favorable aggregates.
The results motivate studying how opposing programs change, with explicit intervention authority, behavior-linked feedback, and historical-opponent regression.

%===============================================================================

\clearpage
% The acknowledgments are automatically included only in the final and preprint versions of the paper.
% \acknowledgments{}

%===============================================================================

\bibliography{references}  % .bib

%===============================================================================

% \clearpage
% \appendix
% \input{appendix}

\end{document}